\documentclass[10pt,twocolumn,letterpaper]{article}

\usepackage{iccv}
\usepackage{times}
\usepackage{graphicx}
\usepackage{amsmath}
\usepackage{amssymb}
\usepackage{booktabs}

\usepackage{blindtext}
\usepackage{multirow}
\usepackage{bm}
\usepackage{bbm} 
\usepackage{comment}
\usepackage{algorithm}
\usepackage{algorithmic}
\usepackage{caption}
\usepackage{subcaption}
\usepackage{array} 
\usepackage{booktabs}
\usepackage[dvipsnames]{xcolor}
\usepackage[safe]{tipa} 
\usepackage{color, colortbl}
\usepackage[misc,geometry]{ifsym} 
\usepackage{multirow, makecell}
\makeatletter
\newcommand\figcaption{\def\@captype{figure}\caption}
  \newcommand\tabcaption{\def\@captype{table}\caption}
\renewcommand\paragraph{
  \@startsection{paragraph} 
  {4} 
  {\z@} 
  {.5em \@plus1ex \@minus.2ex} 
  {-1.5em} 
  {\normalfont\normalsize\bfseries} 
}
\makeatother
\newlength\savewidth\newcommand\shline{\noalign{\global\savewidth\arrayrulewidth
  \global\arrayrulewidth 1pt}\hline\noalign{\global\arrayrulewidth\savewidth}}
\makeatletter
\def\@fnsymbol#1{\ensuremath{\ifcase#1\or \dagger\or \ddagger\or
   \mathsection\or \mathparagraph\or \|\or **\or \dagger\dagger
   \or \ddagger\ddagger \else\@ctrerr\fi}}
\makeatother

\newcommand{\tableCellHeight}{1}
\newcommand{\tabstyle}[1]{
  \setlength{\tabcolsep}{#1}
  \renewcommand{\arraystretch}{\tableCellHeight}
  \centering
  \small
}

\definecolor{tabhighlight}{HTML}{e5e5e5}
\definecolor{citecolor}{HTML}{0071bc}

\usepackage[pagebackref,breaklinks,colorlinks]{hyperref}
\usepackage{multirow,multicol}
\usepackage[accsupp]{axessibility}
\usepackage[capitalize]{cleveref}
\crefname{section}{Sec.}{Secs.}
\Crefname{section}{Section}{Sections}
\Crefname{table}{Table}{Tables}
\crefname{table}{Tab.}{Tabs.}

\iccvfinalcopy 

\def\iccvPaperID{4706} 

\ificcvfinal\fi

\begin{document}

\title{LoGoPrompt: Synthetic Text Images Can Be Good Visual Prompts for Vision-Language Models}
\author{\textbf{
Cheng Shi, 
Sibei Yang\thanks{Corresponding author}}\\ 
School of Information Science and Technology, ShanghaiTech University \\
{\tt\small \{shicheng2022,yangsb\}@shanghaitech.edu.cn}\\
\small{Project Page:} \href{https://chengshiest.github.io/logo/}{\small{https://chengshiest.github.io/logo}} \small{$\dagger$: Corresponding author}\\
}

\twocolumn[{%
  \renewcommand\twocolumn[1][]{#1}%
  \maketitle
    \vspace{-20pt}
    \captionsetup{type=figure}
    \centering
    \includegraphics[width=\textwidth]{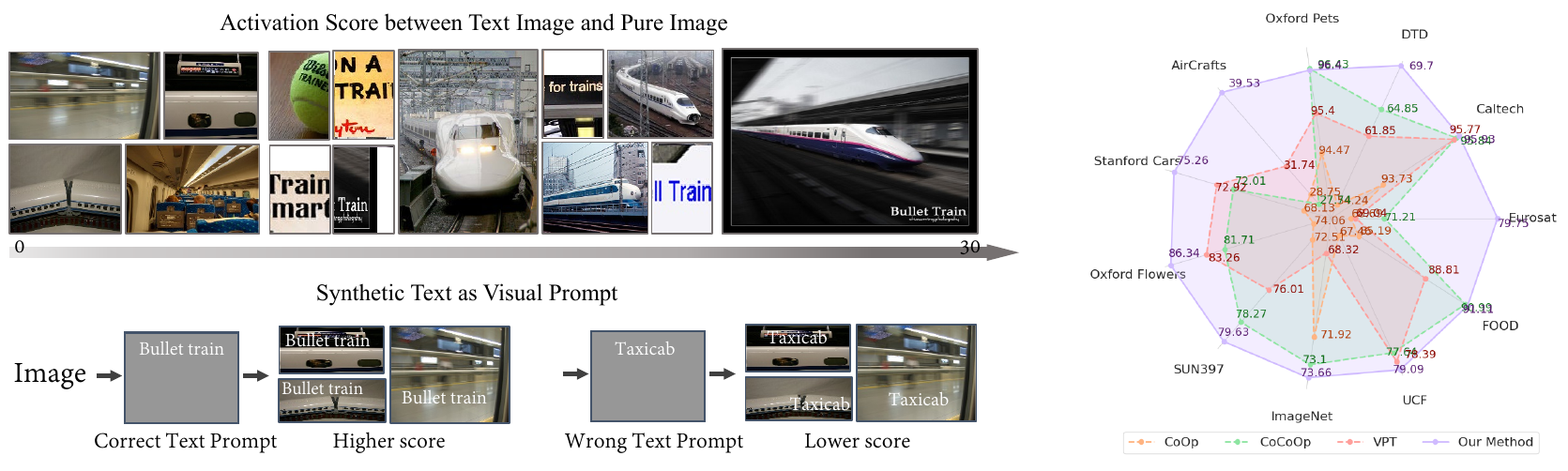}
    \vspace{-12pt}
    \caption{Why can synthetic text images be good visual prompts for CLIP: (a) Images are sorted in ascending order of their zero-shot CLIP scores with the text \textit{``A photo of train"}. (b) Significant improvement of zero-shot CLIP scores for images with synthetic text images as visual prompts. (c) Comparison of base-to-new generalization among prompt-tuning methods.}
    \label{fig:intro}
    \vspace{14pt}
}]

\ificcvfinal\thispagestyle{empty}\fi

\begin{abstract}
Prompt engineering is a powerful tool used to enhance the performance of pre-trained models on downstream tasks. For example, providing the prompt ``Let's think step by step" improved GPT-$3$'s reasoning accuracy to $63$\% on MutiArith while prompting ``a photo of" filled with a class name enables CLIP to achieve $80$\% zero-shot accuracy on ImageNet. 
While previous research has explored prompt learning for the visual modality, analyzing what constitutes a good visual prompt specifically for image recognition is limited. In addition, existing visual prompt tuning methods' generalization ability is worse than text-only prompting tuning. 
This paper explores our key insight: synthetic text images are good visual prompts for vision-language models! To achieve that,  we propose our LoGoPrompt, which reformulates the classification objective to the visual prompt selection and addresses the chicken-and-egg challenge of first adding synthetic text images as class-wise visual prompts or predicting the class first. Without any trainable visual prompt parameters, experimental results on 16 datasets demonstrate that our method consistently outperforms state-of-the-art methods in few-shot learning, base-to-new generalization, and domain generalization. 
\\

\end{abstract}
\vspace{-1cm}
\section{Introduction}
\label{sec:intro}

Large-scale contrastive vision-language models (VLMs) like CLIP~\cite{radford2021learning} and ALIGN~\cite{jia2021scaling}, pretrained on large-scale image-text pairs via contrastive learning, encode general knowledge about the alignment of visual concepts and textual sequences. VLMs with dual-encoder separately encode images and texts into vectors in joint embedding space, enabling the transfer for downstream image classification by treating image vectors as ``image features" and text vectors corresponding to classes as ``classification weights", respectively. The class-specific weights can encoded from handcraft \textit{prompts}~\cite{radford2021learning, jia2021scaling, furst2021cloob, li2021supervision, liu2021pre}, such as ``a photo of a $[\textit{class}]$'', where the $[\textit{class}]$ token is filled with real class names. Thanks to exploring open-set visual concepts with high-capacity text encoder, VLMs with handcraft prompts show impressive potential in zero-shot transfer to image classification tasks.

Recently, CoOp~\cite{zhou2021coop} and CoCoOp~\cite{zhou2022conditional} utilize prompt tuning~\cite{lester2021power, liu2021gpt, liu2021pre} to learn a continuous text prompt with only a few shots of images and improve zero-shot VLMs' performance. 
However, the text prompt tuning can only change the ``classification weights" but keeps the ``image features" unchanged. It is a sub-optimal solution for adaption, making it hard to classify images with different classes but close image features. 
Therefore, some works~\cite{xing2022class, bahng2022visual, jia2022visual} introduce visual prompts to VLMs for simultaneously adjusting the ``image features" and ``classification weights". The visual prompts of VPT~\cite{jia2022visual} and DPT~\cite{xing2022class} are specific to ViT architecture~\cite{vit}, and they adapt to downstream tasks by tuning additional image patch token embeddings. 
Considering the generality to different families of visual backbones, the recent work~\cite{bahng2022visual} proposes to learn image perturbation to perform data-space adaption. Although it explores a new visual prompting paradigm, the performance improvement is limited, especially in the same few-shot setting as the above-mentioned methods. Furthermore, these visual prompts~\cite{xing2022class, bahng2022visual, jia2022visual} seem to affect VLMs' generalization ability, whose base-to-new generalization performance is lower than zero-shot VLMs and approaches with text-only prompting~\cite{zhou2021coop, zhou2022conditional} (see {Figure~\ref{fig:intro}). 

Therefore, we propose to explore visual prompting for VLMs' adaption, which can (1) work for different backbone families such as CNNs and Transformers, (2) effectively adapt to downstream classification tasks with few shots of images, and (3) preserve the VLMs' generalization ability. 
To achieve these goals, we follow~\cite{bahng2022visual} to perform data-space adaption (\ie, adapting on image pixels) and then analyze how VLMs understand different images, especially the classification accuracy gaps between images with the same class. We are surprised to observe that the images with class name text have very high confidence to classify to the class, even only lower than the simplest cases, as shown in Figure~\ref{fig:intro}. The empirical study~\cite{goh2021multimodal} about multimodal neurons of CLIP validates our observation. For a class, both images with the class name text and generic natural images of that class can activate the same neurons important for classifying that class. 

The observation motivates our key insight: we can use synthetic images with class name text as the visual prompts for VLMs! 
However, it is not trivial to effectively utilize synthetic images because simply treating the synthetic images as extra image data cannot achieve consistent performance improvement across different datasets and with different image shots. 
In this paper, we propose to use synthetic images with class name text (\ie, the class-wise visual prompts) to modify images in the training set so that class-wise synthetic parts can help VLMs perceive class-relevant content in the original images for classification. In the training phase, given a training image and its ground-truth class, we can easily transform it with its class-wise visual prompt, such as randomly replacing one of the original image's pixel blocks with the synthetic one, as shown in Figure~\ref{fig:intro}. However, in the testing stage, since the test image's class is unknown, it remains unclear which class-wise visual prompt should be used to benefit the classification prediction of the test image, which is a chicken-and-egg problem. Therefore, we reformulate the downstream classification objective as the visual prompt selection: select the synthetic images of the correct class for the original image as its visual prompts. Specifically, for a training image, we transform it into multiple images with different class-wise visual prompts and maximize the similarity of the image with the ground-truth visual prompt to the text features of that ground-truth class while minimizing the similarities of other images.

Furthermore, to develop the visual prompt selection to be effective and efficient, we propose a min-max contrastive learning objective and introduce hard negative mining~\cite{facenet,defense}. The min-max contrastive learning first groups an original image and its transformed image with the class-wise visual prompt as a group. 
Then, it maximizes the minimal similarity to the text features in one ground-truth group while minimizing the maximal similarity in negative groups to preserve the ability to classify the original image. 

Despite the simplicity, our novel insights of using synthetic images with class name text as visual prompts and visual prompt selection learning strategy are particularly effective. 
Without requiring any tuning of visual prompts, our method significantly outperforms state-
of-the-art methods~\cite{zhou2022conditional} on base-to-new generalization, especially compared to previous visual prompting approaches~\cite{xing2022class, bahng2022visual, jia2022visual}, as shown in Figure~\ref{fig:intro}. Notice that our visual prompts have no parameters to learn, but we still have text prompts for tuning following existing works~\cite{zhou2021coop,xing2022class,zhou2022conditional}. To verify whether our method can benefit from visual prompt tuning, we extend a tuning version for visual prompts, further boosting the performance on few-shot classification. 
We name our proposed method LoGoPrompt, since we incorporate synthetic text onto images in a manner akin to the application of logos onto visuals. 
We evaluate and compare LoGoPrompt with state-of-the-art methods on 16 datasets that cover diverse image classification tasks. 
In summary, we make the following contributions:
\begin{itemize}
\setlength{\itemsep}{0pt}
\setlength{\parsep}{0pt}
\setlength{\parskip}{0pt}
        \item We are the first to propose using synthetic images with the text class name as visual prompts for VLMs. Our visual prompts
        can work for different backbone families, \eg, CNNs and Transformers.
        \item We reformulate the classification objective to visual prompt selection to address the chicken-and-egg challenge of adding class-wise visual prompts and achieve the selection via min-max contrastive learning.
	\item Despite the simplicity, experiments on 16 datasets demonstrate our novel insight and learning strategy particularly effectively, consistently outperforming state-of-the-art methods in base-to-new generalization, few-shot learning, and domain generalization. 
\end{itemize}

\section{Related Work}

\noindent\textbf{Vision-Language Pre-training Models}, pre-trained on image-text corpora for modeling vision and language, have shown great potential for transferable to downstream vision and language tasks. The pre-training approaches mainly contain the BERT-like masked-language and masked-region modeling methods~\cite{lu2019vilbert, su2019vl, tan2019lxmert, chen2020uniter}, contrastive learning for learning a joint embedding space of vision and language~\cite{radford2021learning, jia2021scaling, li2021align, zhai2022lit}, and vision-language multimodal autoregressive techniques~\cite{cho2021unifying, ramesh2021zero}. In this paper, we focus on the contrastive vision-language models (VLMs) that adopt a dual encoder to encode images and texts into the joint embedding space and use contrastive learning to align the visual and textual representations. The VLMs, particularly CLIP~\cite{radford2021learning} and ALIGN~\cite{jia2021scaling}, leverage hundreds of millions and even billions of image-text pairs to learn transferable visual representation from textual supervision and shows impressive zero-shot performance for various image classification tasks. 
CLIPPO~\cite{Tschannen_2023_CVPR} also treats render images as text to achieve similar results compared with CLIP without a text-specific tower or embedding. 
In this work, we devise our LoGoPrompt based on the CLIP.

\noindent\textbf{Data-efficient Fine-tuning for VLMs} learns to adapt the VLMs to downstream tasks with a few shots of samples. Following~\cite{zhou2021coop, zhou2022conditional}, we focus on the image classification task. Linear Probe~\cite{radford2021learning} straightforwardly fixes the visual backbone and learns a classifier for classification. CLIP-Adapter~\cite{gao2021clip} and Tip-Adapter~\cite{zhang2021tip} introduce additional feature adapters to improve the few-shot classification of CLIP. Unlike the feature adapters, prompt tuning methods~\cite{zhou2021coop, zhou2022conditional, zhu2022prompt} learn input prompts to keep a closer objective form to the pre-training task and can achieve better generalization ability. CoOp~\cite{zhou2021coop} proposes a context optimization method to learn task-specific text prompts. CoCoOp~\cite{zhou2022conditional} extends the static text prompts of CoOp by learning image-conditional dynamic prompts to improve the base-to-new generalization. ProGrad~\cite{zhu2022prompt} addresses the issue of forgetting general knowledge by regularizing the tuning step not to conflict with the zero-shot CLIP's prediction. 
ProDA~\cite{Lu_2022_CVPR} prompt distribution learning to handle the varying visual representation.
In this work, we focus on prompt-based tuning and propose to take synthetic images with class name text as visual prompts to improve both base-to-new generalization and few-shot classification.

\noindent\textbf{Visual Prompt Tuning.} Inspired by the success of prompt tuning in the NLP community~\cite{brown2020language, li2021prefix, liu2021p}, recent works~\cite{jia2022visual, yao2021cpt, bahng2022visual, wen2022visual, sohn2022visual} explore visual prompt tuning for vision and vision-language models. CPT~\cite{yao2021cpt} learns color-based prompts to adapt the BERT-like VLMs to the visual grounding task. VPT~\cite{jia2022visual} and VPT for generative transfer~\cite{sohn2022visual} learn visual tokens for Transformer-based vision models for better transfer to visual recognition and image synthesis tasks, respectively. Unlike adapting to computer vision tasks, VPT for text classification~\cite{wen2022visual} deploys the VLMs in text classification via visual prompt tuning. 
The recently published DPT~\cite{xing2022class} and the unpublished VP~\cite{bahng2022visual} are the most relevant works to ours, and both of them adapt the contrastive VLMs to downstream image classification. DPT extends VPT by dynamically generating visual prompts via the cross-attention between text prompts features and visual image patch token embeddings. But, its visual prompts are specific to Transformer-based visual backbones. In contrast, VP learns the visual prompts in pixel space for generalizable to different backbones. However, it cannot obtain satisfied few-shot performance, demonstrating that simply learning pixel-space perturbations for images cannot achieve data-efficient tuning for VLMs. 
Unlike previous methods, our class-wise visual prompts are synthetic images with the class name text, which naturally work for different backbone families and have no trainable parameters, and can improve state-of-the-art prompt tuning for VLMs. 
\section{Method}

Figure~\ref{fig:framwork} shows an overview of our proposed LoGoPrompt. We first briefly review the CLIP~\cite{radford2021learning} and CoOp~\cite{zhou2021coop} in Section~\ref{subsec:p}. Next, we introduce LoGoPrompt, including the visual prompt generation, min-max contrastive learning, and extension to tunable visual prompts, in Section~\ref{subsec:STprompt}. Like CoOp, our LoGoPrompt is applicable to other CLIP-like contrastive VLMs.

\begin{figure*}[t]
\begin{center}
\includegraphics[width=1\linewidth]{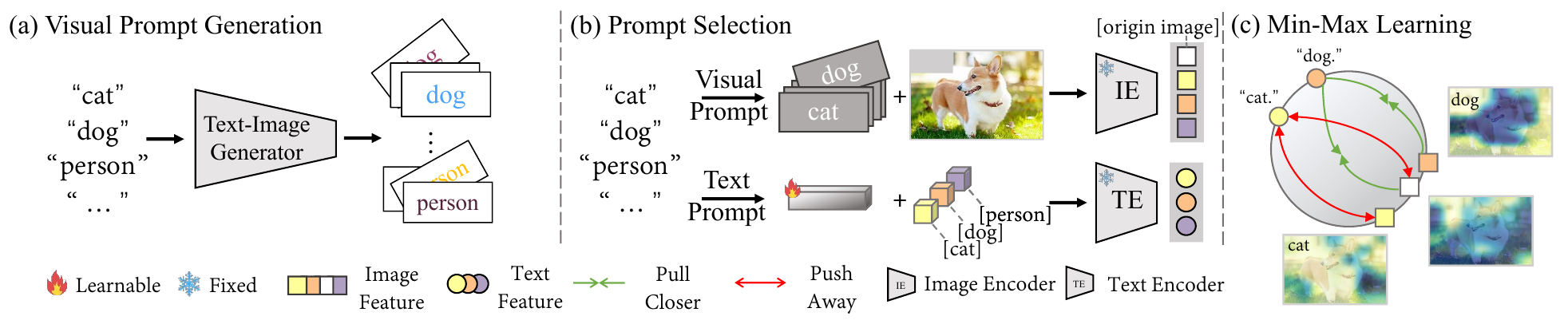}
\end{center}
\vspace{-0.5cm}
\caption{\textbf{Overview of LoGoPrompt}, 
which (a) generates class-wise visual prompts as synthetic images with text class names and (b) reformulates the classification objective to visual prompt selection to address the chicken-and-egg challenge by (c) the proposed min-max contrastive learning. 
}
\label{fig:framwork}
\vspace{-0.2cm}
\end{figure*}
\subsection{Preliminary}
\label{subsec:p}
\noindent\textbf{Contrastive Language-Image Pre-training (CLIP)}~\cite{radford2021learning} has a pair of image and text encoders, where the image encoder $f(\cdot)$ can be either a CNN (e.g., ResNet~\cite{he2016deep}) or a Vision Transformer (e.g., ViT~\cite{vit}), and the text encoder $g(\cdot)$ is a Transformer~\cite{vaswani2017attention}. During training, CLIP adopts the dual encoder to separately encode images and text into vectors in joint embedding space and utilizes contrastive loss to maximize the cosine similarities of the real image-text vector pairs while minimizing the cosine similarities of incorrect pairs. After being pre-trained on highly-diversified hundreds of millions of image-text pairs, CLIP is available for computing the text-image similarity and can be generalized to downstream image recognition without fine-tuning. 

Let $\boldsymbol{x}$ be an input image and $\{[class]_c\}_{c=1}^C$ be the $C$ categories for classification, where $[class]_c$ represents the class name of the $c$-th class. With a handcraft prompt like ``a photo of a $[class]_c$.", the prediction probabilities are:
\begin{equation}
	p(\hat{y}=c|\boldsymbol{x}) = \frac{\exp(\cos(f(\boldsymbol{x}), g(\boldsymbol{l}_c))/\tau)}{\sum_{i=1}^{C}{\exp(\cos(f(\boldsymbol{x}), g(\boldsymbol{l}_i))/\tau)}}
\end{equation} 
where $\boldsymbol{l}_c$ is the sequence embeddings of the handcraft prompt for $c$-th class, $\tau$ is a learned temperature coefficient, $\cos(\cdot,\cdot)$ represents the cosine similarity, and $f(\cdot)$ and $g(\cdot)$ are the image encoder and text encoder, respectively. 

\label{subsec:coop}
\noindent\textbf{Context Optimization (CoOp)}~\cite{zhou2021coop} addresses time-consuming and unstable issues of prompt engineering by replacing the fixed handcraft prompt with the tunable prompt that can be learned from data. The tunable prompt is composed of $M$ learnable continues context vectors $\boldsymbol{u}=[\boldsymbol{u}_1, \boldsymbol{u}_2, ..., \boldsymbol{u}_M]$ that has same dimension as the word embedding and the word embedding of class names. For $c$-th class, the prompt $\boldsymbol{p}_c$ is $[\boldsymbol{u}_1, \boldsymbol{u}_2, ..., \boldsymbol{u}_M, \boldsymbol{e}_c]$, where $\boldsymbol{e}_c$ is the word embedding of the $c\text{-th}$ class's class name $[class]_c$. CoOp optimizes the context vectors $\boldsymbol{u}$ by minimizing the cross-entropy loss between the ground truth and prediction probability as follows,
 \begin{equation}
 	\begin{aligned}
 	p(\hat{y}=c|\boldsymbol{x}) &= \frac{\exp(\cos(f(\boldsymbol{x}), g(\boldsymbol{p}_c))/\tau)}{\sum_{i=1}^{C}{\exp(\cos(f(\boldsymbol{x}), g(\boldsymbol{p}_i))/\tau)}}, \\
 	\mathcal{L}_{\text{ce}} &= -\log{p(\hat{y}=y|\boldsymbol{x})}\\
 	\end{aligned}
 \end{equation} 
where $y$-th class is the ground-truth class of the image $\boldsymbol{x}$. Note that only the context vectors $\boldsymbol{u}$ are updated during the tuning, while CLIP's image encoder $f(\cdot)$ and the text encoder $g(\cdot)$ are frozen. We also follow the same training protocol as CoOp.

\subsection{Synthetic Text Images Can Be Good Visual Prompts for VLMs}
\label{subsec:STprompt}
We now present the proposed LoGoPrompt. As illustrated in Figure~\ref{fig:framwork}, our LoGoPrompt first generates visual prompts as synthetic images with class name texts (Section~\ref{subsubsec:csp}). Then, it reformulates the learning objective of image classification to the visual prompt selection by utilizing min-max contrastive learning (Section~\ref{subsubsec:cpl}). Moreover, LoGoPrompt can easily extend the frozen visual prompts to be tunable to boost the performance further (Section~\ref{subsubsec:gcspl}).

\vspace{-0.2cm}
\subsubsection{Visual Prompt Generation}\label{subsubsec:csp}
Motivated by the observation that images with class name text easily activate the same classification neurons as general natural images of the same class (presented in the Introduction~\ref{sec:intro}), we propose to use synthetic images with class name text as visual prompts. As the class name text is independent and different for different classes, the visual prompts are naturally class-wise. The class-wise visual prompts are expected to help VLMs perceive class-relevant content in general natural images for better classification. 

\noindent\textbf{Class-wise Visual Prompts} are defined as class-specific text images, \ie, the synthetic pixel blocks of the text classes rendered on the empty background, where the colors of the text classes and the background are randomly generated. 
Figure~\ref{fig:framwork}\textcolor{red}{a} shows class-wise visual prompts of two different categories of dog: ``Saluki" and ``Otterhound".
Formally, for the $c$-th class, we denote its visual prompt as $V_{c}\in\mathbb{R}^{h \times w \times 3}$, where the $h$ and $w$ are the height and width of the pixel block, respectively. The class-wise visual prompts $\{V_c\}_{c=1}^C$ have the same size of $h \times w$ across different classes. 

\noindent\textbf{Class-conditional Images.} Given an input image $\boldsymbol{x}$, we use the class-wise visual prompt $V_{c}$ to transform it into a class-conditional image $\boldsymbol{x}_c$ of the $c$-th class. There are different ways to perform the transform, and we simply randomly replace a $h \times w$ pixel block of the image $\boldsymbol{x}$ with the $c$-th class's visual prompt $V_{c}$ for simplicity. 
The class-conditional image $\boldsymbol{x}_c$ can be treated as an augmented image of the original image $\boldsymbol{x}$ on the specific class $c$. For example, as shown in Figure~\ref{fig:framwork}\textcolor{red}{c}, the visual prompt of ``dog” enhances the original image with the class ``dog".    
\subsubsection{Min-Max Contrastive Learning for Visual Prompt Selection}
\label{subsubsec:cpl}

\noindent\textbf{Problem Formulation: Visual Prompt Selection.} Although class-conditional images enhance the original images by applying class-wise visual prompts, it remains unclear how a class-wise visual prompt should be selected to enhance the image in the test phase. As the test image's class is unknown, it becomes a chicken-and-egg problem: should the class of the test image be predicted first to obtain the class-wise visual prompt or should the test image be augmented with the class-wise visual prompt for better prediction? 
To overcome the challenge, we reformulate the classification objective as the visual prompt selection: learn to select candidate class-wise visual prompts for the image during the training phase so that the same selection strategy can be applied in the test phase. 

Specifically, we construct the real and negative class-conditional images for a training image by applying ground truth and other class-wise visual prompts, respectively. Then, we utilize contrastive loss to maximize the prediction probability for the real image while minimizing that of the negative image. More importantly, to ensure that original images can be classified, we construct the groups of class-conditional and original images and improve the traditional contrastive loss to min-max one for applicable to groups of images.

\noindent\textbf{Sample Construction for Synthetic and Original Image-Class Pairs} constructs real and negative image-class pairs for class-conditional and original images. Given the input image $\boldsymbol{x}$ and its ground-truth class $y$, we construct a group of real image-class pairs and $K$ groups of negative image-class pairs. Each group has two image-class pairs, one for the original image and one for the class-conditional image. Our construction rules are as follows:
\begin{itemize}
\setlength{\itemsep}{0pt}
\setlength{\parsep}{0pt}
\setlength{\parskip}{0pt}
    \item  Group of real image-class pairs is $[(\boldsymbol{x}, y)^{+}, (\boldsymbol{x}_y, y)^{+}]$. The reason is that the ground-truth label $y$ is the correct class for both the original image $\boldsymbol{x}$ and the class-conditional image $\boldsymbol{x}_y$ of the class $y$. For example, the pairs of (image, ground-truth class ``dog") and (image with visual prompt ``dog", ground-truth class ``dog") are correct, as shown in Figure~\ref{fig:framwork}\textcolor{red}{c}. 
    \item The $K$ groups of negative image-text pairs are $\{[(\boldsymbol{x}, c_k)^{-}, (\boldsymbol{x}_{c_k}, c_k)^{-}]\}_{k=1}^K$, where class $c_k \neq y$. The class $c_k$ is incorrect class for original image $\boldsymbol{x}$ and all class-conditional images $\{\boldsymbol{x}_c\}_{c=1}^C$, including the class-conditional image $\boldsymbol{x}_{c_k}$ of class $c_k$. Since class $y$ is the correct class for image $\boldsymbol{x}$, the visual prompt $V_{c_k}$ of class $c_k$ cannot activate visual concepts relevant to class $c_k$ from the image $\boldsymbol{x}$ of class $y$, resulting in $(\boldsymbol{x}_{c_k}, c_k)^{-}$ is a negative pair. The class-wise visual prompts are expected to help VLMs to perceive class-relevant visual concepts in the image and should not change the original inherent visual semantics of the image. For example, an image of ``a dog" should not be recognized as ``a cat" by adding ``cat" visual prompt. 
\end{itemize}

\noindent\textbf{Min-Max Contrastive Loss} maximizes the similarity of real pairs $[(\boldsymbol{x}, y)^{+}, (\boldsymbol{x}_y, y)^{+}]$ while minimizing that of negative pairs $\{[(\boldsymbol{x}, c_k)^{-}, (\boldsymbol{x}_{c_k}, c_k)^{-}]\}_{k=1}^K$ to optimize the context vectors $\boldsymbol{u}$ of text prompts $\{\boldsymbol{p}_i\}_{i=1}^C$ (see CoOp in Section~\ref{subsec:coop}). Our visual prompts are synthetic images with class name text, which have no parameters to learn. Therefore, we only need to optimize continuous context vectors $\boldsymbol{u}$ following CoOp. 
Formally, we extend the InfoNCE loss~\cite{oord2018representation} on the groups of real and negative pairs as follows,
\begin{equation}
	\begin{aligned}
			&p(\hat{y}=c|\boldsymbol{x}) = \frac{\exp(\cos(f(\boldsymbol{x}), g(\boldsymbol{p}_c))/\tau)}{\sum_{i=1}^{C}{\exp(\cos(f(\boldsymbol{x}), g(\boldsymbol{p}_i))/\tau)}}, \\
			&\mathcal{L}_{\text{N}} = -\log{\frac{\min(p(y|\boldsymbol{x}), p(y|\boldsymbol{x}_y))}{\sum_{k=1}^K\max(p(c_k|\boldsymbol{x}), p(c_k|\boldsymbol{x}_{c_k}))}} \\
	\end{aligned}
\end{equation}
where the $\min(\cdot,\cdot)$ and $\max(\cdot,\cdot)$ operations obtain the minimum and maximum matching probabilities for the real and negative groups, respectively. Note that the min-max contrastive loss maximizes the minimal probability within the real group while minimizing the maximal probability within negative groups to preserve the ability to classify the original image.   

During tuning, we adopt the hard negative mining to select the $K$ classes $\{c_k\}_{k=1}^K$ with the top-$K$ prediction probabilities $p(\hat{y}=c_k|\boldsymbol{x})$ for the original image $\boldsymbol{x}$ other than the ground-truth class $y$. During inference, we also first obtain the top-$K$ prediction classes for the original image, and then select the class with the highest prediction probability for the original or the corresponding class-conditional image as the predicted class, \ie, $\hat{y}=\text{argmax}_{{c_k}_{k=1}^K}\max(p(c_k|\boldsymbol{x}), p(c_k|\boldsymbol{x}_{c_k}))$.

\subsubsection{Extension to Tunable Prompts} \label{subsubsec:gcspl}
\noindent\textbf{Visual Prompt:} We use the synthetic pixel blocks of the text classes to initialize the visual prompts before tuning. Next, we optimize both the visual prompts $\{V_c\}_{c=1}^C$ and textual context vectors $\boldsymbol{u}$ during the tuning via the proposed contrastive prompt learning in Section~\ref{subsubsec:cpl}.

\noindent\textbf{Text Prompt:} Following~\cite{zhou2021coop}, we adopt a class-specific text prompt and optimized it in a two-stage strategy. In the first stage, inspired by the Meta-Net of CoCoOp~\cite{zhou2022conditional}, we use a lightweight MLP to directly generate class-specific text prompts from handcraft prompts' text vectors extracted by CLIP's text encoder. In the second stage, we fine-tune the learned prompt from stage one directly on seen classes.

\begin{table*}[t]
    \setlength{\belowcaptionskip}{7pt}
    \tabstyle{5.5pt}
        \caption{\textbf{Accuracy comparison for base-to-new generalization setting}. The prompts are learned from the base classes (16 shots) for prompt-based tuning methods). Following CoCoOp, ViT-B/16 of CLIP is used as the vision backbone. H: Harmonic mean~\cite{xian2017zero}. 
        The latter three methods employ visual prompts.
        The superior performance of LoGoPrompt on both base and new classes shows its strong generalizability. 
        }
    \label{tab:results_generalization}
    \begin{subtable}[t]{.23\textwidth}
    \tabcolsep=0.22em
    \centering
    \caption{\textbf{Average over 11 datasets}.}
    \vspace{-0.1cm}
    \begin{tabular}{l cc|c}
    
    \toprule
    & Base & New & H \\
    \midrule
    CLIP & 69.34 & 74.22 & 71.70 \\
    CoOp & 82.69 & 63.22 & 71.66 \\
    CoCoOp & 80.47 & 71.69 & 75.83 \\
    VPT & 80.81 & 70.36 & 74.68 \\
    DPT & 84.18 & 66.47 & 74.28 \\
    \rowcolor{tabhighlight}
    Ours & \textbf{84.47} & \textbf{74.24} & \textbf{79.03} \\
    \bottomrule
    \end{tabular}
    \end{subtable}
    ~
    \begin{subtable}[t]{.23\textwidth}
    \tabcolsep=0.22em
    \centering
    \caption{ImageNet.}
    \vspace{-0.1cm}
    \begin{tabular}{l cc|c}
    \toprule
    & Base & New & H \\
    \midrule
    CLIP & 72.43 & 68.14 & 70.22 \\
    CoOp & 76.47 & 67.88 & 71.92\\
    CoCoOp & 75.98 & 70.43 & 73.10 \\
    VPT & 70.93 & 65.90 & 68.32 \\
    DPT & \textbf{76.95} & 68.14 & 72.28 \\
    \rowcolor{tabhighlight}
    Ours & 76.74 & \textbf{70.83} & \textbf{73.66} \\
    \bottomrule
    \end{tabular}
    \end{subtable}
    ~
    \begin{subtable}[t]{.23\textwidth}
    \tabcolsep=0.22em
    \centering
    \caption{Caltech101.}
    \vspace{-0.1cm}
    \begin{tabular}{l cc|c}
    \toprule
    & Base & New & H \\
    \midrule
    CLIP & 96.84 & \textbf{94.00} & 95.40 \\
    CoOp & 98.00 & 89.81 & 93.73 \\
    CoCoOp & 97.96 & 93.81 & 95.84 \\
    VPT & 97.86 & 93.76 & 95.77 \\
    DPT & \textbf{98.49} & 92.36 & 95.33 \\
    \rowcolor{tabhighlight}
    Ours & 98.19 & 93.78 & \textbf{95.93} \\
    \bottomrule
    \end{tabular}
    \end{subtable}
    ~
    \begin{subtable}[t]{.23\textwidth}
    \tabcolsep=0.22em
    \centering
    \caption{OxfordPets.}
    \vspace{-0.1cm}
    \begin{tabular}{l cc|c}
    \toprule
    & Base & New & H \\
    \midrule
    CLIP & 91.17 & 97.26 & 94.12 \\
    CoOp & 93.67 & 95.29 & 94.47 \\
     CoCoOp & 95.20 & \textbf{97.69} & \textbf{96.43} \\
     VPT & 94.81 & 96.00 & 95.40 \\
    DPT & 95.07 & 95.69 & 95.38 \\
    \rowcolor{tabhighlight}
    Ours & \textbf{96.07} & 96.31 & 96.18 \\
    \bottomrule
    \end{tabular}
    \end{subtable}
    ~
    \begin{subtable}[t]{.23\textwidth}
    \tabcolsep=0.22em
    \centering
    \caption{StanfordCars.}
    \vspace{-0.1cm}
    \begin{tabular}{l cc|c}
    \toprule
    & Base & New & H \\
    \midrule
    CLIP & 63.37 & \textbf{74.89} & 68.65 \\
    CoOp & 78.12 & 60.40 & 68.13 \\
    CoCoOp & 70.49 & 73.59 & 72.01 \\
    VPT & 72.46 & 73.38 & 72.92 \\
    DPT & \textbf{82.07} & 60.72 & 69.80 \\
    \rowcolor{tabhighlight}
    Ours & 78.36 & 72.39 & \textbf{75.26} \\
    \bottomrule
    \end{tabular}
    \end{subtable}
    ~
    \begin{subtable}[t]{.23\textwidth}
    \tabcolsep=0.22em
    \centering
    \caption{Flowers102.}
    \vspace{-0.1cm}
    \begin{tabular}{l cc|c}
    \toprule
    & Base & New & H \\
    \midrule
    CLIP & 72.08 & \textbf{77.80} & 74.83 \\
    CoOp & 97.60 & 59.67 & 74.06 \\
    CoCoOp & 94.87 & 71.75 & 81.71 \\
    VPT & 95.39 & 73.87 & 83.26 \\
    DPT & 98.13 & 64.14 & 77.58 \\
    \rowcolor{tabhighlight}
    Ours & \textbf{99.05} & 76.52 & \textbf{86.34} \\
    \bottomrule
    \end{tabular}
    \end{subtable}
    ~
    \begin{subtable}[t]{.23\textwidth}
    \tabcolsep=0.22em
    \centering
    \caption{Food101.}
    \vspace{-0.1cm}
    \begin{tabular}{l cc|c}
    \toprule
    & Base & New & H \\
    \midrule
    CLIP & 90.10 & 91.22 & 90.66 \\
    CoOp & 88.33 & 82.26 & 85.19 \\
    CoCoOp & 90.70 & 91.29 & 90.99 \\
    VPT & 89.88 & 87.76 & 88.81 \\
    DPT & 88.41 & 85.71 & 87.04 \\
    \rowcolor{tabhighlight}
    Ours & \textbf{90.82} & \textbf{91.41} & \textbf{91.11} \\
    \bottomrule
    \end{tabular}
    \end{subtable}
    ~
    \begin{subtable}[t]{.23\textwidth}
    \tabcolsep=0.22em
    \centering
    \caption{FGVCAircraft.}
    \vspace{-0.1cm}
    \begin{tabular}{l cc|c}
    \toprule
    & Base & New & H \\
    \midrule
    CLIP & 27.19 & \textbf{36.29} & 31.09 \\
    CoOp & 40.44 & 22.30 & 28.75 \\
    CoCoOp & 33.41 & 23.71 & 27.74 \\
    VPT & 33.10 & 30.49 & 31.74 \\
    DPT & 43.98 & 25.83 & 32.55 \\
    \rowcolor{tabhighlight}
    Ours & \textbf{45.98} & 34.67 & \textbf{39.53} \\
    \bottomrule
    \end{tabular}
    \end{subtable}
    ~
    \begin{subtable}[t]{.23\textwidth}
    \tabcolsep=0.22em
    \centering
    \caption{SUN397.}
    \vspace{-0.1cm}
    \begin{tabular}{l cc|c}
    \toprule
    & Base & New & H \\
    \midrule
    CLIP & 69.36 & 75.35 & 72.23 \\
    CoOp & 80.60 & 65.89 & 72.51 \\
    CoCoOp & 79.74 & 76.86 & 78.27\\
    VPT & 79.66 & 72.68 & 76.01 \\
    DPT & 80.38 & 65.72 & 72.31 \\
    \rowcolor{tabhighlight}
    Ours & \textbf{81.20} & \textbf{78.12} & \textbf{79.63} \\
    \bottomrule
    \end{tabular}
    \end{subtable}
    ~
    \begin{subtable}[t]{.23\textwidth}
    \tabcolsep=0.22em
    \centering
    \caption{DTD.}
    \vspace{-0.1cm}
    \begin{tabular}{l cc|c}
    \toprule
    & Base & New & H \\
    \midrule
    CLIP & 53.24 & 59.90 & 56.37 \\
    CoOp & 79.44 & 41.18 & 54.24 \\
    CoCoOp & 77.01 & 56.00 & 64.85 \\
    VPT & 79.15 & 50.76 & 61.85 \\
    DPT & \textbf{83.49} & 49.48 & 62.10 \\
    \rowcolor{tabhighlight}
    Ours & 82.87 & \textbf{60.14} & \textbf{69.70} \\
    \bottomrule
    \end{tabular}
    \end{subtable}
    ~
    \begin{subtable}[t]{.23\textwidth}
    \tabcolsep=0.22em
    \centering
    \caption{EuroSAT.}
    \vspace{-0.1cm}
    \begin{tabular}{l cc|c}
    \toprule
    & Base & New & H \\
    \midrule
    CLIP & 56.48 & 64.05 & 60.03 \\
    CoOp & 92.19 & 54.74 & 68.69 \\
    CoCoOp & 87.49 & 60.04 & 71.21 \\
    VPT & 93.01 & 54.89 & 69.04 \\
    DPT & 92.96 & 58.35 & 71.70 \\
    \rowcolor{tabhighlight}
    Ours & \textbf{93.67} & \textbf{69.44} & \textbf{79.75} \\
    \bottomrule
    \end{tabular}
    \end{subtable}
    ~
    \begin{subtable}[t]{.23\textwidth}
    \tabcolsep=0.22em
    \centering
    \caption{UCF101.}
    \vspace{-0.1cm}
    \begin{tabular}{l cc|c}
    \toprule
    & Base & New & H \\
    \midrule
    CLIP & 70.53 & \textbf{77.50} & 73.85 \\
    CoOp & 84.69 & 56.05 & 67.46 \\
    CoCoOp & 82.33 & 73.45 & 77.64 \\
    VPT & 82.67 & 74.54 & 78.39 \\
    DPT & 86.02 & 64.99 & 74.04 \\
    \rowcolor{tabhighlight}
    Ours & \textbf{86.19} & 73.07 & \textbf{79.09} \\
    \bottomrule
    \end{tabular}
    \end{subtable}
\vspace{-0.2cm}
\end{table*}

\begin{figure*}[ht]
    \centering
\begin{minipage}[h]{1\linewidth}
    \begin{minipage}[t]{0.247\linewidth}
    \includegraphics[width=1\linewidth]{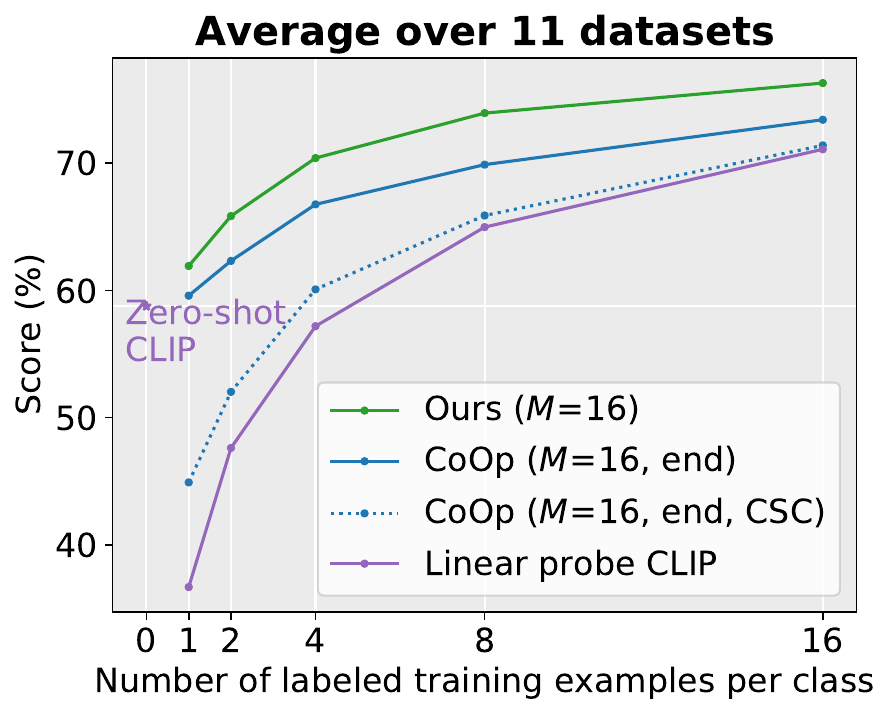}
    \end{minipage}
    \begin{minipage}[t]{0.247\linewidth}
    \includegraphics[width=1\linewidth]{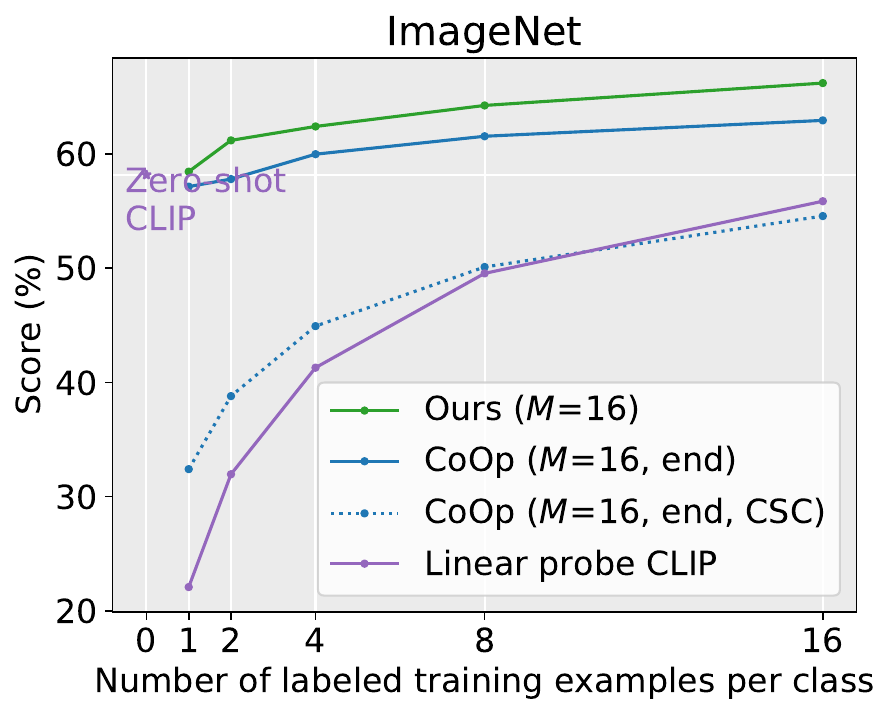}
    \end{minipage}
    \begin{minipage}[t]{0.247\linewidth}
    \includegraphics[width=1\linewidth]{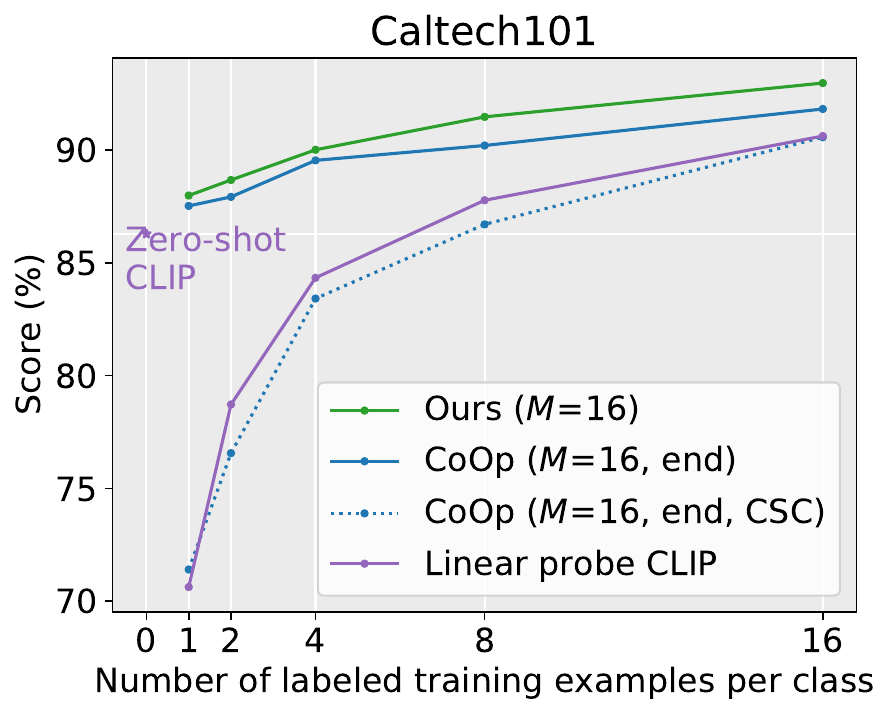}
    \end{minipage}
    \begin{minipage}[t]{0.247\linewidth}
    \includegraphics[width=1\linewidth]{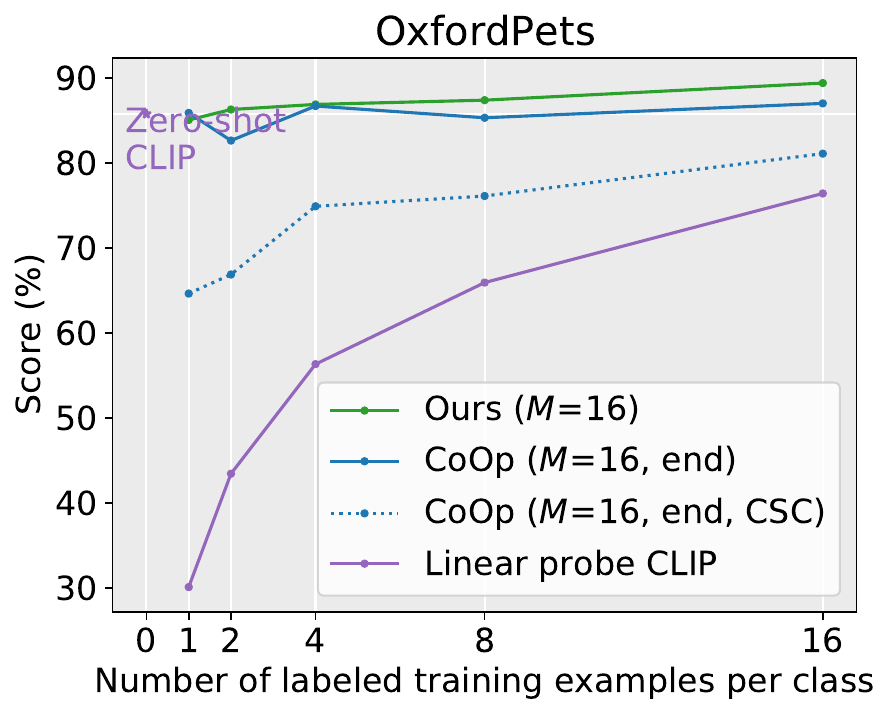}
    \end{minipage}
    
    \end{minipage}
    
    \begin{minipage}[h]{1\linewidth}
    \begin{minipage}[t]{0.247\linewidth}
    \includegraphics[width=1\linewidth]{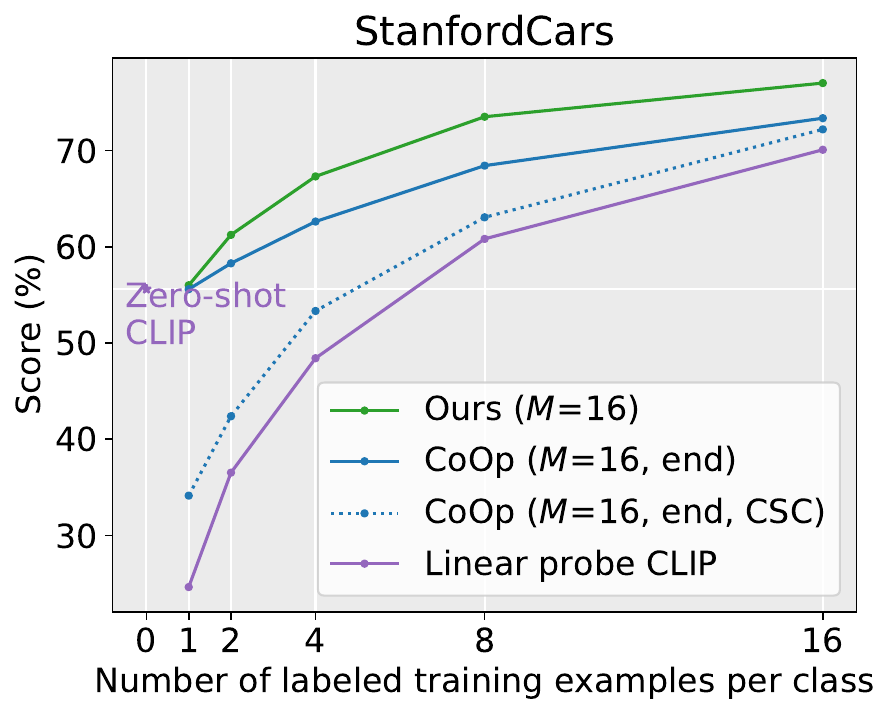}
    \end{minipage}
    \begin{minipage}[t]{0.247\linewidth}
    \includegraphics[width=1\linewidth]{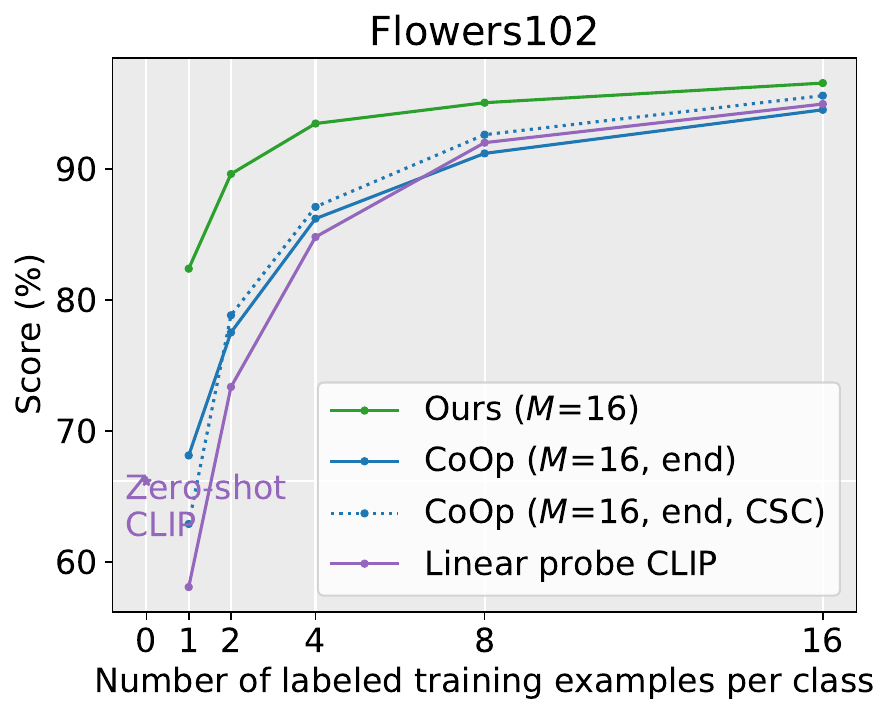}
    \end{minipage}
    \begin{minipage}[t]{0.247\linewidth}    
    \includegraphics[width=1\linewidth]{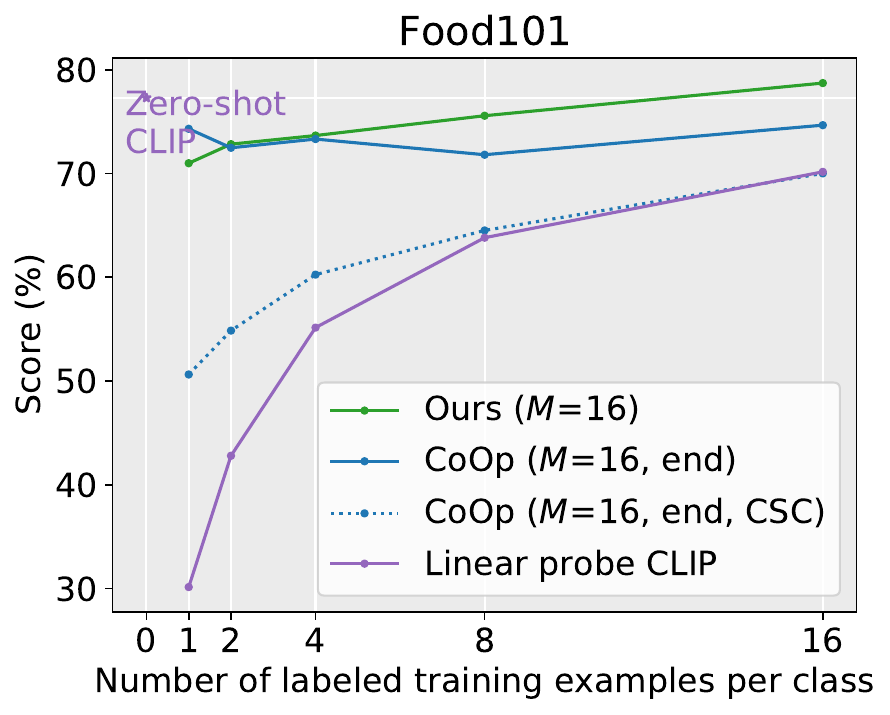}
    \end{minipage}
    \begin{minipage}[t]{0.247\linewidth}
    \includegraphics[width=1\linewidth]{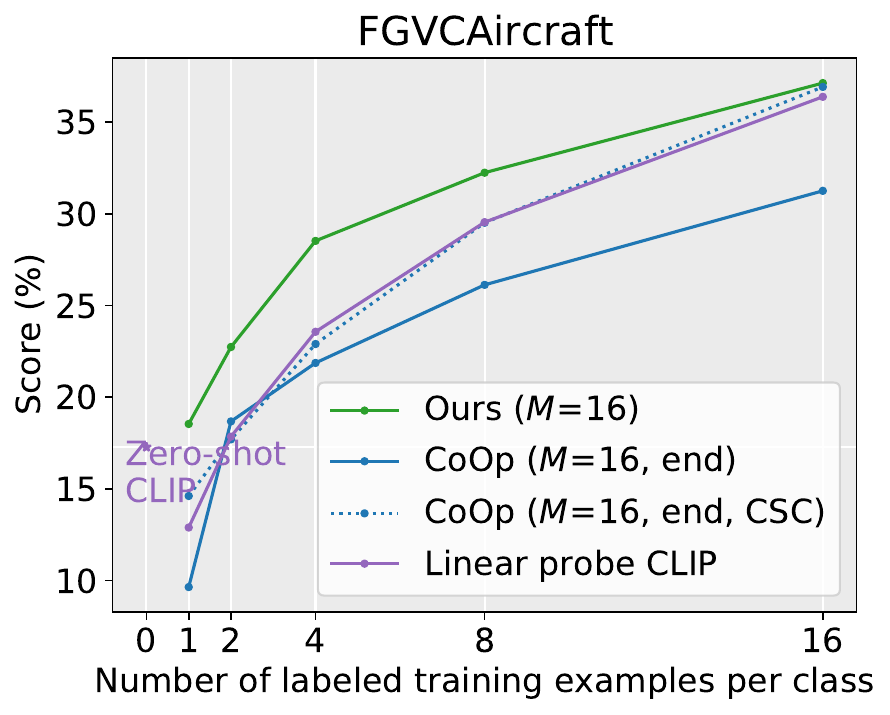}
    \end{minipage}
    \end{minipage}
    
    \begin{minipage}[h]{1\linewidth}
    \begin{minipage}[t]{0.247\linewidth}
    \includegraphics[width=1\linewidth]{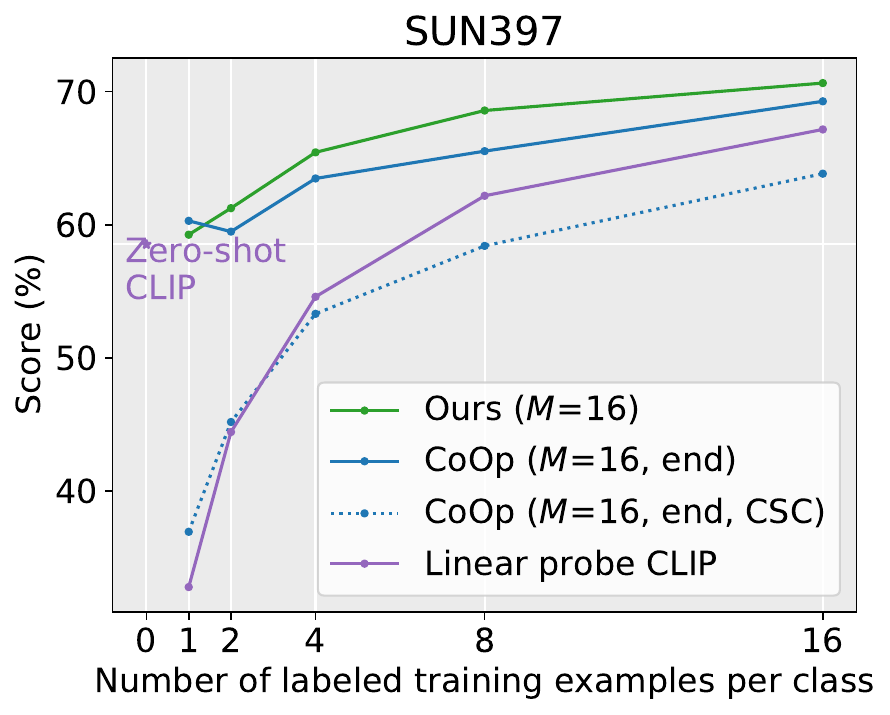}
    \end{minipage}
    \begin{minipage}[t]{0.247\linewidth}    
    \includegraphics[width=1\linewidth]{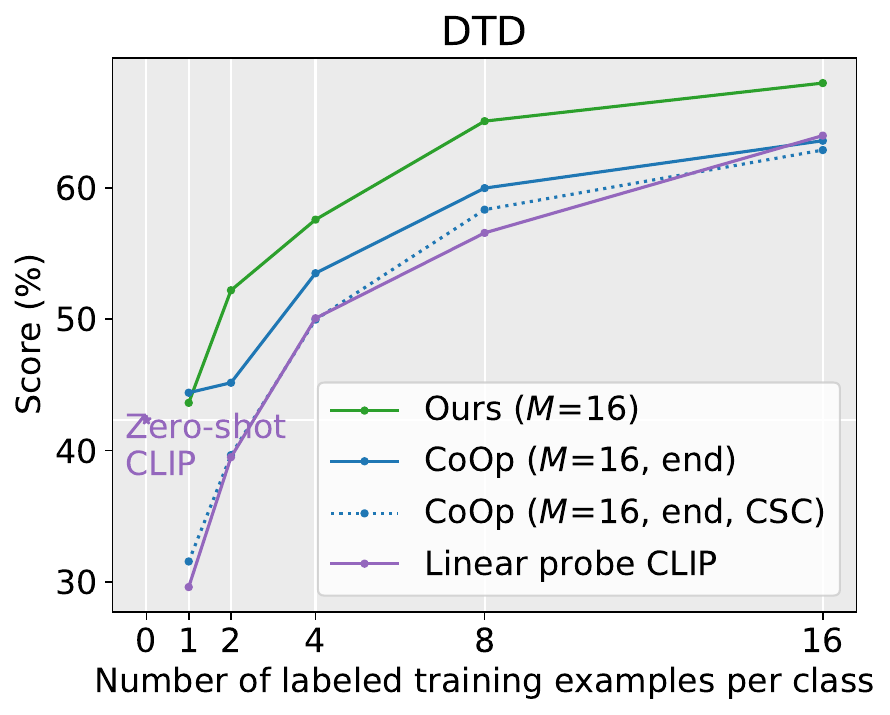}
    \end{minipage}
    \begin{minipage}[t]{0.247\linewidth}
    \includegraphics[width=1\linewidth]{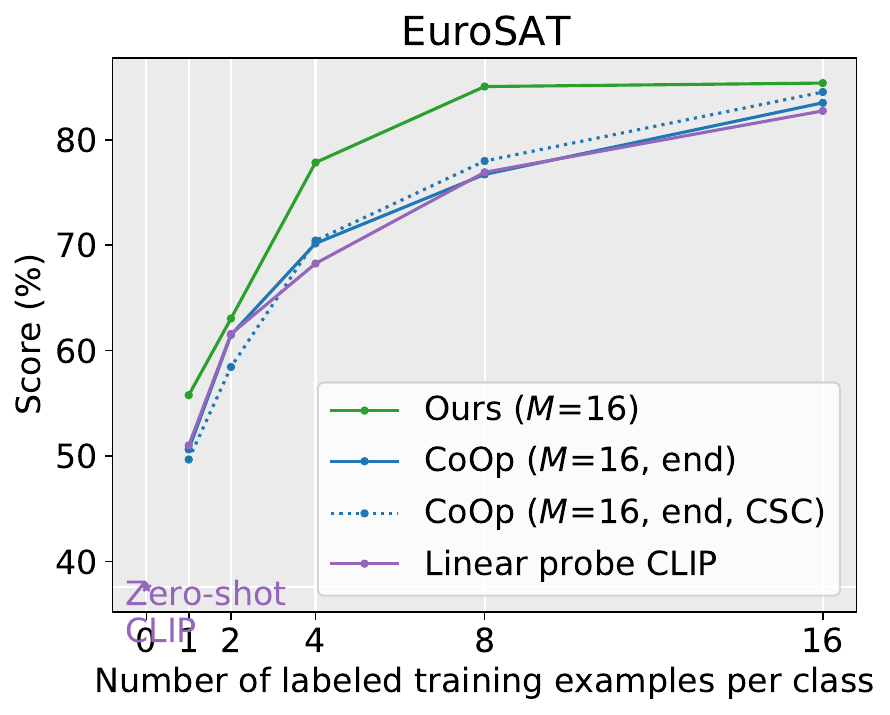}
    \end{minipage}
    \begin{minipage}[t]{0.247\linewidth}
    \includegraphics[width=1\linewidth]{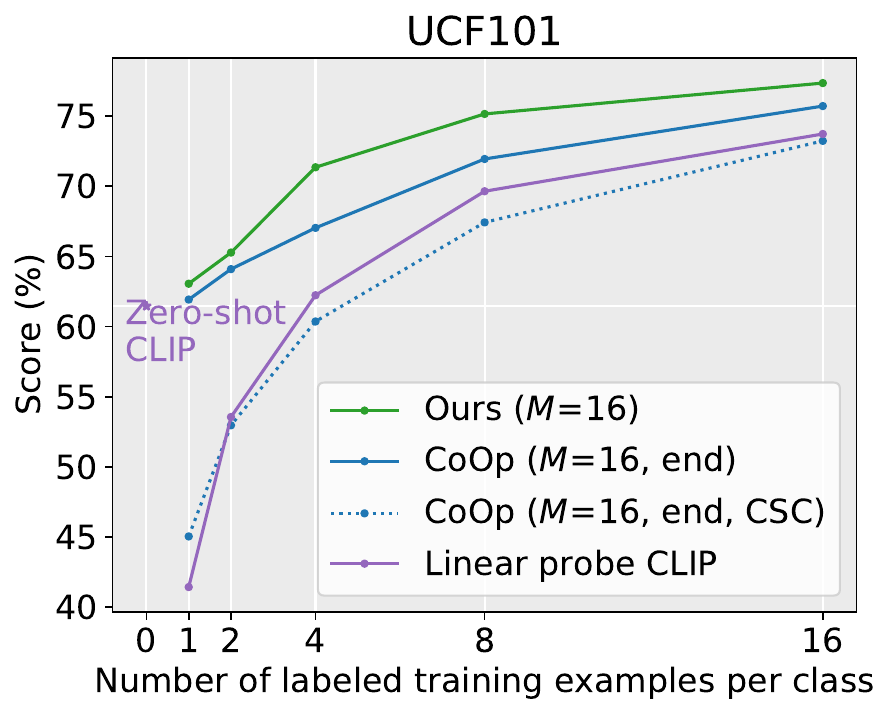}
    \end{minipage}
    \end{minipage}
    \caption{Accuracy comparison of few-shot classification. LoGoPrompt consistently outperforms compared methods on all the 11 datasets. Following CoOp, ResNet-50 of CLIP is used as the vision backbone.}
    \label{fig:my_label}
\end{figure*}

\section{Experiments}
We evaluate the proposed LoGoPrompt on three problem settings: (1) generalization from base classes to new classes (Section~\ref{subsec:btn}), (2) few-shot classification (Section~\ref{subsec:fs}), (3) domain generalization (Section~\ref{subsec:dg}). 


\noindent\textbf{Datasets.} For the first two settings, we follow CLIP~\cite{radford2021learning} and CoOp~\cite{zhou2021coop} to evaluate on 11 image classification datasets, \ie, ImageNet (Img)~\cite{deng2009imagenet}, Caltech101 (Cal)~\cite{fei2004learning}, OxfordPets (Pet)~\cite{parkhi2012cats}, StanfordCars (Car)~\cite{krause20133d}, Flowers102 (Flo)~\cite{nilsback2008automated}, Food101 (Foo)~\cite{bossard2014food}, FGVCAircraft (FGV)~\cite{maji2013fine}, SUN397 (SUN)~\cite{xiao2010sun}, DTD~\cite{cimpoi2014describing}, EuroSAT (Eur)~\cite{helber2019eurosat}, and UCF101 (UCF)~\cite{soomro2012ucf101}. For domain generalization, we adopt ImageNet~\cite{deng2009imagenet} as the source and ImageNetV2 (V2)~\cite{recht2019imagenet}, ImageNet-Sketch (Sketch)~\cite{wang2019learning}, ImageNet-A (A)~\cite{hendrycks2021natural} and ImageNet-R (R)~\cite{hendrycks2021many} as target datasets following CoCoOp~\cite{zhou2022conditional}. 


\noindent\textbf{Training details.} For all three problem settings, we follow the experimental settings of CoOp~\cite{zhou2021coop} and CoCoOp~\cite{zhou2022conditional} for a fair comparison, including dataset splits, data augmentation, training schedule, shots of samples, backbones, length of context tokens (\ie, $M$ is 16 for few-shot classification and 4 for other problems), \etc. 
The $K$ is set to 5 for all the experiments. Please refer to the Supplementary for more details.

\subsection{Base-to-New Generalization} \label{subsec:btn}
To evaluate the base-to-new generalization ability, we follow CoCoOp to train LoGoPrompt on 16 images per base class in the training set and report the performance on both base and new classes of the test set. As shown in Table~\ref{tab:results_generalization}, the average performance of LoGoPrompt consistently surpasses other methods in terms of all three metrics, demonstrating that our LoGoPrompt has not only impressive few-shot learning ability but also has strong generalizability. 

Specifically, the Harmonic mean~\cite{xian2017zero} indicates the generalization trade-off between base classes and new classes. 
We evaluate DPT~\cite{xing2022class}\footnote{The results of DPT are borrowed directly from its original authors, and we extend our appreciation for their contribution of outcomes.} and VPT based on their official implementations. Compared with DPT and VPT that also use visual prompts, LoGoPrompt shows impressive generalization ability, which improves the average accuracy on new classes by $3.88$\%. 
Moreover, LoGoPrompt shows a clear gain of $3.20\%$ over the previous best-performing CoCoOp and surpasses CoCoOp on 10 out of 11 datasets. For the accuracy on base classes, LoGoPrompt outperforms the strong few-shot baseline CoOp by nearly $2\%$ on average accuracy and beats all the methods on all the datasets. For the accuracy of unseen new classes, LoGoPrompt improves the average accuracy of learning-based CoOp and CoCoOp by $11.02\%$ and $2.55\%$, respectively. Moreover, LoGoPrompt even achieves the average accuracy of zero-shot CLIP and outperforms CLIP by $2.69\%$, $2.77\%$ and $5.39\%$ on ImageNet, SUN397 and EuroSAT datasets, respectively. 
The results demonstrate that our class-specific visual and text prompts tuning method does not hurt the generalization ability of CLIP even if it learns the prompts based on the base classes during tuning.

\subsection{Few-Shot Classification} \label{subsec:fs}
Figure~\ref{fig:my_label} summarizes accuracy ($\%$) comparison with 1, 2, 4, 8 and 16 shots on 11 datasets. Our LoGoPrompt consistently outperforms prompt-tuning models and zero-shot CLIP~\cite{radford2021learning} on all the datasets with different shots, which demonstrates LoGoPrompt's data effectiveness and generalization ability on various types of datasets. Specifically, our LoGoPrompt outperforms CoOp~\cite{zhou2021coop} by $3.28\%$ and $3.58\%$ on average accuracy given 2 and 4 shots, respectively. The results show that LoGoPrompt can achieve superior performance even with limited training samples. Moreover, LoGoPrompt improves the average accuracy over all the shots by $7.91\%$, $6.32\%$, and $3.04\%$ on Flowers102, FGVCAircraft and StanfordCars datasets respectively, expressing its effectiveness on fine-grained image recognition.

Although we focus on the line of prompt-based tuning methods, we also compare LoGoPrompt to other fine-tuning approaches, \ie, Linear Probe~\cite{radford2021learning}, CLIP-Adapter~\cite{gao2021clip} and Tip-Adapter~\cite{zhang2021tip}. The results illustrated in Table~\ref{tab:sota} show that LoGoPrompt outperforms other fine-tuning methods on various datasets. Besides, our method can benefit from the feature adapter. With an adapter, the performance of our LoGoPrompt further improves and surpasses state-of-the-art methods on all the datasets. Note that we add a feature adapter following Tip-Adapter but do not search for the optimal hyper-parameter on thousands of sets of hyper-parameters like Tip-Adapter. 

Moreover, we evaluate the few-shot performance of Visual Prompting~\cite{bahng2022visual}, the most relevant visual prompting work to ours, using the officially released code. As shown in Table~\ref{tab:sota}, \textcolor{black}{our LoGoPrompt surpasses Visual Prompting~\cite{bahng2022visual} by large margins, which demonstrates the effectiveness of our class-specific visual prompt tuning.} Please refer to the Supplementary for detailed results and more comparisons. 

\begin{table*}[t]
    \tabstyle{3pt}
    \caption{\textbf{Comparison of LoGoPrompt, CoOp and other fine-tuning methods in the few-shot classification setting} (16 shots). Following CoOp, ResNet-50 of CLIP is used as the vision backbone for all the models. The performance of LoGoPrompt surpasses both prompt-based learning and other fine-tuning methods. \textcolor{black}{VP and TP are abbreviation for Visual prompt and Text prompt. Best results are bold.}}
    \tabcolsep=0.08em
    \label{tab:sota}
    \begin{tabular}{l| cc | ccccccccccc|c}
    \toprule
    & Prompt & Adapter
    & Ima~\cite{deng2009imagenet} & Cal~\cite{fei2004learning} & Pet~\cite{parkhi2012cats} & Car~\cite{krause20133d} & Flo~\cite{nilsback2008automated} & Foo~\cite{bossard2014food} & FGV~\cite{maji2013fine} & SUN~\cite{xiao2010sun} & DTD~\cite{cimpoi2014describing} & Eur~\cite{helber2019eurosat} & UCF~\cite{soomro2012ucf101} & \emph{Average} \\
    \midrule
    CoOp~\cite{zhou2021coop} &TP && 62.95 & 91.83 & 87.01 & 73.36 & 94.51 & 74.67 & 31.26 & 69.26 & 63.58 & 83.53 & 75.71 & 73.42 \\
    Vis Prompt~\cite{bahng2022visual} & VP+TP && 50.65 & 77.02 & 69.64 & 56.99 & 89.97 & 59.99 & 23.01 & 57.01 & 55.67 & 73.19 & 67.11 & 61.84 \\
    Ours &VP+TP && \textbf{66.34} & \textbf{92.98} & \textbf{89.40} & \textbf{77.01} & \textbf{96.55} & \textbf{78.73} & \textbf{37.14} & \textbf{70.63} & \textbf{67.97} & \textbf{85.40} & \textbf{77.35} & \textbf{76.31\tiny{$\pm$ 0.52}} \\
    \midrule
    
    Tip~\cite{zhang2021tip} & TP &\checkmark& 62.01 & 90.18 & 88.14 & 66.77 & 89.89 & 77.83 & 29.76 & 66.85 & 60.93 & 70.54 & 70.58 & 70.32 \\
    Tip-F~\cite{zhang2021tip} & TP &\checkmark& 65.51 & 92.86 & 89.70 & 75.74 & 94.80 & 79.43 & 35.55 & 71.47 & 66.55 & 84.54 & 78.03 & 75.83 \\
    
    Ours &VP+TP &\checkmark& \textbf{67.34} & \textbf{93.23} & \textbf{90.35} & \textbf{77.64} & \textbf{96.83} & \textbf{79.77} & \textbf{38.70} & \textbf{72.13} & \textbf{70.07} & \textbf{85.74} & \textbf{78.32} & \textbf{77.28\tiny{$\pm$ 0.43}}  \\
    \bottomrule
    \end{tabular}
    \vspace{-0.2cm}
\end{table*}

\begin{table}[t]
    \tabstyle{3.0pt}
    \begin{tabular}{l ccccc c}
    \toprule
    \tabcolsep=0.005em
    & Source & \multicolumn{4}{c}{Target} & Avg \\ \cmidrule(lr){2-2} \cmidrule(lr){3-6} \cmidrule(lr){7-7}& ImageNet & V2 & Sketch & A & R  \\ 
    \midrule
    CLIP~\cite{radford2021learning} & 66.73 & 60.83 & 46.15 & 47.77 & 73.96 & 59.08 \\
    CoOp~\cite{zhou2021coop}  & 71.51 & 64.20 & 47.99 & 49.71 & 75.21 & 61.72 \\
    CoCoOp~\cite{zhou2022conditional}  & 71.02 & 64.07 & 48.75 & 50.63 & 76.18 & 62.13\\
    VPT~\cite{xing2022class}  & 70.72 & 58.22 & 44.67 & 43.00 & 71.86 & 57.69\\
    DPT~\cite{xing2022class}  & 72.38 & 64.96 & 47.46 & 45.63 & 74.81 & 61.05\\
    Ours & \textbf{75.27} & \textbf{66.65} & \textbf{48.99} & \textbf{51.36} & \textbf{76.85} & \textbf{63.82}\\
    \bottomrule
    \end{tabular}
    \caption{\textbf{Accuracy comparison of domain generalization}. Learning-based methods are trained on ImageNet with 1,000 classes and 16 images per class. Following CoCoOp, ViT-B/16 of CLIP is used as the vision backbone. LoGoPrompt is more domain-generalizable than others.
    }
    \label{tab:dg}
\vspace{-0.2cm}
\end{table}

\subsection{Domain Generalization} \label{subsec:dg}
Following CoOp and CoCoOp, we validate the generalization of our LoGoPrompt to out-of-distribution data. We evaluate the accuracy by transferring LoGoPrompt trained on ImageNet to four target benchmarks, and the results are shown in Table~\ref{tab:dg}. Our LoGoPrompt consistently outperforms compared models on all four target datasets, and the average improvement on the source and target datasets are $2.10\%$ and $1.69\%$ compared to CoOp and CoCoOp, respectively. Furthermore, since the instance-conditional prompt learning of CoCoOp cannot be optimized in parallel, it requires dozens of times training time than ours. 
The results give evidence that our visual and text prompts are more domain-generalizable and efficient. We also evaluate visual prompting methods, \ie VPT and DPT. And our method significantly outperforms them by $6.13\%$. 

\subsection{Further Analysis}

\begin{table}[t]
\tabcolsep=0.15em
\small
\begin{center}
\begin{tabular}{lc|lc}

\shline
\multicolumn{1}{l}{\bf \textit{(a) Visual-Pro Type}} & &  \multicolumn{1}{l}{\bf \textit{(b) Visual-Pro Size}}&\\

text augmentation   & 63.12  & ratio = 1/14   & 66.02\\
paired-text-image  & 63.54  & ratio = 1/7   & \textbf{66.34}\\
text-image-selection   & \textbf{66.34} & ratio = 2/7   & 66.17\\

\hline
\multicolumn{1}{l}{\bf \textit{(c) Text-Pro Length}} & & \multicolumn{1}{l}{\bf \textit{(d) Visual-Pro Location}} &\\
$M$ = 2   & 65.58  & top   & 65.98\\
$M$ = 8  & 66.14  & bottom   & 66.01\\
$M$ = 16  & \textbf{66.34}  & rand   & \textbf{66.34\tiny{$\pm$ 0.21}}\\

\hline
\end{tabular}
\end{center}
\vspace{-14pt}
\caption{\textbf{Further analysis and ablation study on ImageNet dataset,} which reports the few-shot (16-shot) accuracy with ResNet-50 as the vision backbone.
    }
\vspace{-16pt}
\label{tab:abla_a}
\end{table}

\noindent\textbf{Visual Prompt.} The variants of the visual prompt shown in Table~\ref{tab:abla_a}\textcolor{red}{a} contains: 
(1) By directly treating synthetic text images as additional one-shot training images, the model's accuracy exhibits slight changes.
(2) By replacing the text augmentation with the paired text-image, which means the standard classification loss is utilized, the model's accuracy achieves a mild gain by $0.42\%$. Notice that paired text-image is also another solution for the chicken-and-egg problem. Despite the performance improvement from (1), it yields only marginal improvements compared with CoOp~\cite{zhou2021coop}. (3) We further introduce our visual prompt selection strategy and min-max contrastive loss. The model's accuracy further improves to $66.34\%$. The results demonstrate not only the success and effectiveness of our strategy but also that it is non-trivial to leverage synthetic images to adapt VLM to downstream tasks better.

Besides, Table~{\ref{tab:abla_a}}\textcolor{red}{b} and \textcolor{red}{\ref{tab:abla_a}d} illustrate the effects of the size and location of the visual prompt. 
Table~{\ref{tab:abla_a}}\textcolor{red}{c} illustrates the effects of the length of the text prompt.
We observe that visual prompts that are too large would dominate the class-conditioned images and ignore the original image information, which will lead to a $0.17$\% performance drop.
Our random location strategy on the location of the visual prompt mitigates the overfitting problem and improves $0.33$\% compared with the fixed location.

\section{Conclusion}
To summarize, we present a new method that employs synthetic images with text class names as visual prompts for VLMs. 
The proposed LoGoPrompt reformulates the classification objective through min-max contrastive learning, overcoming the challenge of adding class-specific visual prompts. Experiment results on 16 datasets demonstrate the efficacy of our approach. 

\noindent\textbf{Acknowledgment:} This work was supported by the National Natural Science Foundation of China (No.62206174), Shanghai Pujiang Program (No.21PJ1410900), Shanghai Frontiers Science Center of Human-centered Artificial Intelligence (ShangHAI), MoE Key Laboratory of Intelligent Perception and Human-Machine Collaboration (ShanghaiTech University), and Shanghai Engineering Research Center of Intelligent Vision and Imaging.

{\small
\bibliographystyle{ieee_fullname}
\bibliography{egbib}
}

\end{document}